\documentclass{article} 
\usepackage{iclr2026_conference}
\usepackage{times}

\usepackage{amsmath,amsfonts,bm}

\def\1{\bm{1}}

\DeclareMathAlphabet{\mathsfit}{\encodingdefault}{\sfdefault}{m}{sl}
\SetMathAlphabet{\mathsfit}{bold}{\encodingdefault}{\sfdefault}{bx}{n}

\newcommand{\E}{\mathbb{E}}

\usepackage{hyperref}
\hypersetup{
    colorlinks,
    linkcolor={red!50!black},
    citecolor={red!50!black},
    urlcolor={red!80!black}
}
\usepackage{cleveref}
\usepackage{url}
\usepackage{amsthm}
\usepackage{graphicx}
\usepackage{subcaption} 
\usepackage{tabularx} 
\usepackage{booktabs}
\usepackage{multirow}
\usepackage{listings}

\usepackage{graphicx}
\usepackage{enumitem}
\usepackage{tcolorbox}
\tcbuselibrary{listings, skins, breakable}

\newtcolorbox{promptbox}[2][]{enhanced, breakable,
  colback=gray!10, colframe=black!60,
  coltitle=white, fonttitle=\bfseries,
  title=#2, #1}

\newcommand{\Maestro}{\textsc{Maestro}}

\title{\Maestro: Learning to Collaborate via Conditional Listwise Policy Optimization for Multi-Agent LLMs}

\iclrfinalcopy  

\author{%
  \textbf{Wei Yang, Jiacheng Pang, Shixuan Li, Paul Bogdan, Stephen Tu, Jesse Thomason }\\
  University of Southern California \\
  \texttt{\{wyang930,pangj,sli97750,pbogdan,stephen.tu,jessetho\}@usc.edu}\\
}

\begin{document}
\maketitle

\thispagestyle{plain}
\pagestyle{plain}

\begin{abstract}
Multi-agent systems (MAS) built on Large Language Models (LLMs) are being used to approach complex problems and can surpass single model inference. However, their success hinges on navigating a fundamental cognitive tension: the need to balance broad, \textbf{divergent} exploration of the solution space with a principled, \textbf{convergent} synthesis to the optimal solution. Existing paradigms often struggle to manage this duality, leading to premature consensus, error propagation, and a critical credit assignment problem that fails to distinguish between genuine reasoning and superficially plausible arguments. To resolve this core challenge, we propose the \textbf{Multi-Agent Exploration–Synthesis framework Through Role Orchestration (\Maestro{})}, a principled paradigm for collaboration that structurally decouples these cognitive modes. \Maestro{} uses a collective of parallel Execution Agents for diverse exploration and a specialized Central Agent for convergent, evaluative synthesis. To operationalize this critical synthesis phase, we introduce \textbf{Conditional Listwise Policy Optimization (CLPO)}, a reinforcement learning objective that disentangles signals for strategic decisions and tactical rationales. By combining decision-focused policy gradients with a list-wise ranking loss over justifications, CLPO achieves clean credit assignment and stronger comparative supervision. Experiments on mathematical reasoning and general problem-solving benchmarks demonstrate that \Maestro{}, coupled with CLPO, consistently outperforms existing state-of-the-art multi-agent approaches, delivering absolute accuracy gains of 6\% on average and up to 10\% at best. 
\end{abstract}


\section{Introduction}

The rise of large language models (LLMs) have enabled a new type of \emph{multi-agent system} (MAS) ~\citep{park2023generative,chen2023autoagents, zhu2025lamarl}, where multiple model instances collaborate to tackle problems that exceed the capacity of any single model~\citep{zhang2024proagent,qiao2024autoact,han2025joyagents}. By distributing roles and enabling structured interaction, MASs hold the promise of achieving robustness, creativity, and reliability that emerge from collective intelligence~\citep{cheng2024exploring,pezeshkpour2024reasoning}. 
At the heart of any effective collaborative system lies a fundamental cognitive tension. Early work in the psychology of creativity~\citep{runco1995cognition,brophy2001comparing,zhang2020metacontrol} emphasizes that intelligent problem-solving requires a dynamic balance between two seemingly contradictory modes of thought: \textbf{Divergent Creativity} and \textbf{Convergent Critique}. Guilford's theory of divergent and convergent thinking~\citep{guilford1967nature} formalizes this duality: divergence is the generative process of exploring a wide array of alternative hypotheses, while convergence is the evaluative process of comparing, refining, and synthesizing these options.
Without the former, a system risks premature closure; without the latter, it risks incoherence and indecision~\citep{sternberg1991investment,cropley2006praise}. 
Achieving a principled and effective synergy between these two capabilities is the essential challenge for effective LLM agent collaboration.

Despite their diversity, the limitations of existing paradigms point to a set of recurring requirements for advancing multi-agent collaboration. First, an effective system should strike a balance between \textbf{divergent exploration and convergent synthesis}, ensuring that creativity is not stifled by premature agreement yet also not lost in unbounded search. Second, it should enable \textbf{disentangled credit assignment} across structured outputs~\citep{li2025multi,he2025enhancing}, so that strategic decisions and supporting rationales receive distinct and targeted learning signals rather than being conflated into a single monolithic reward. Third, a robust framework requires \textbf{transparent and scalable interaction protocols}~\citep{qian2024scaling,hu2024scalable}, where information is propagated in analyzable ways that remain efficient as the number of agents and rounds increases~\citep{yang2025agentnet}. Together, these desiderata highlight the limitations of existing approaches and motivate the need for a new paradigm that integrates principled exploration, evaluative precision, and collaborative scalability.

\begin{figure}[t]
    \centering
    \includegraphics[width=1\linewidth]{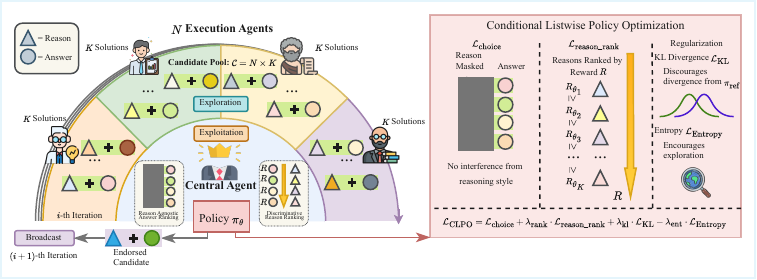}
    \vspace{-24pt}
    \caption{Overview of the \textbf{\Maestro{}} framework. First, $N$ execution agents each generate $K$ candidate reasoning-answer pairs, forming a broad solution pool. A central agent then governs exploitation by applying discriminative selection over the candidate set.
    The decision policy $\pi_\theta$ is trained under \textbf{Conditional Listwise Policy Optimization (CLPO)}, which integrates a choice-aware objective, a reasoning-rank objective, and regularization terms including KL divergence and entropy.} 
    \label{fig:Overall architecture.}
\end{figure}

To address these desiderata, we propose the \textbf{Multi-Agent Exploration–Synthesis framework Through Role Orchestration (\Maestro{})}, a principled paradigm for multi-agent collaboration (Figure~\ref{fig:Overall architecture.}). The effectiveness of \Maestro{} arises not from any single component, but from the synergistic orchestration of specialized roles. \Maestro{} explicitly operationalizes the divergent--convergent duality  through a structured role orchestration: (\emph{i}) \textbf{Divergence as Collective Exploration}, where multiple Execution Agents generate a broad and diverse candidate pool; (\emph{ii}) \textbf{Convergence as List-wise Bayesian Synthesis}, where a Central Agent evaluates these candidates to identify and endorse the most promising solution; and (\emph{iii}) \textbf{Broadcast as Public Conditioning}, where the endorsed solution is propagated back to all agents, guiding the next round of exploration. This cycle of divergence, convergence, and broadcast structures collaboration into 
analyzable and scalable phases.
To further optimize the convergence phase, we introduce \textbf{Conditional Listwise Policy Optimization (CLPO)}, a reinforcement learning objective that disentangles decision-making from rationale generation. Unlike standard GRPO-style sequence-level training, CLPO allocates learning signal separately to \emph{decisions} 
(which candidate to endorse) and \emph{reasons} (why this choice is defensible). \Maestro{} and CLPO constitute a new paradigm for multi-agent collaboration that integrates cognitive inspiration with principled optimization.
The main contributions of this paper are:
\begin{itemize}[nosep,noitemsep,leftmargin=2em]
    \item We introduce the Multi-Agent Exploration–Synthesis framework Through Role Orchestration (\Maestro{}), a principled paradigm for multi-agent collaboration that explicitly operationalizes the divergent--convergent duality through three coordinated phases.
    \item We propose Conditional Listwise Policy Optimization (CLPO), an RL objective that decouples  signals for \emph{decisions} and \emph{reasons}. CLPO combines group-relative decision optimization with listwise rationale ranking for clean credit assignment and stable convergence.
    \item Extensive experiments on mathematical and general reasoning benchmarks show that \Maestro{} with CLPO achieves significant improvements over state-of-the-art baselines.
\end{itemize}

\section{Related Work}
\label{sec:related_work}

We highlight representative related works in multi-agent LLM collaboration and RL for multi-agent LLMs.
For a more in-depth account of related work, see Appendix~\ref{ref:related_work}.

\noindent\textbf{Multi-Agent LLM Collaboration.}
Large language model (LLM) based multi-agent systems have been proposed to overcome the inherent limits of single models in context length, sequential reasoning, and skill breadth~\citep{abdelnabi2023llm,wu2024autogen,yan2025beyond,dai2025multi}. By coordinating multiple agents, these systems can decompose tasks, critique candidate solutions, and integrate diverse perspectives~\citep{hong2023metagpt, chen2023agentverse, qiao2024autoact, pan2024agentcoord}. One common design follows \emph{prestructured coordination}, where communication topologies and protocols are fixed in advance~\citep{chen2024llmarena,mukobi2023welfare,wang2023avalon,abdelnabi2024cooperation}. Debate and peer-review frameworks~\citep{du2023improving,chan2023chateval,liu2024groupdebate} encourage agents to cross-examine one another, while chain or graph structures regulate message flow~\citep{qian2024scaling,liu2023dynamic}. These methods reduce hallucinations and improve consistency but often enforce early convergence, limiting exploration and leaving credit assignment opaque~\citep{hu2024routerbench,yue2025masrouter}. A second line explores \emph{adaptive coordination}, where the collaboration graph is reorganized dynamically during inference. Examples include routing and pruning strategies~\citep{yue2025masrouter,hu2024automated}, as well as workflow and graph-search approaches that optimize interaction structures through reinforcement or evolutionary methods~\citep{zhuge2024gptswarm,zhang2024aflow,zhang2024cut}.

\noindent\textbf{Reinforcement Learning for Multi-Agent LLMs.}
Reinforcement learning (RL) provides a natural mechanism for improving collaboration in multi-agent LLM systems beyond static prompt design~\citep{madaan2023self, zelikman2024quiet, zelikman2022star, zhuang2024yolo,zhu2025lamarl}. Rather than relying solely on prestructured debate or workflow rules, RL enables agents to adapt interaction patterns from feedback, learning when and how to communicate to achieve stronger group performance~\citep{zhou2025reso, wang2024self, xu2025scalable, wan2025rema, park2025maporl,yang2025learning}. These approaches show that reward-driven updates can uncover strategies for dynamic role assignment, coordination, and decision aggregation. A central challenge in this setting is credit assignment~\citep{liu2023forward, zhang2024multi, zhang2024towards, li2024language}. Most existing methods propagate reward at the system level, treating outcomes as global properties of the entire team~\citep{jiang2025qllm, lin2025speaking}. This global reward fails to identify the specific contributions of individual agents or to separate the quality of rationales from the correctness of final decisions. Recent efforts attempt to design more targeted objectives~\citep{wei2025lero, alsadat2024multi}, but principled, fine-grained supervision remains limited. Our work focuses specifically on the convergence step: we view it as a structured optimization problem and design an objective that provides more precise credit assignment than existing system-level rewards.

\section{Methodology}

We introduce a novel learning paradigm to enhance the collective problem-solving capabilities of multi-agent systems. We design a collaborative process through the lens of a new structural framework, the \textbf{Multi-Agent Exploration–Synthesis (\Maestro{})} paradigm, which orchestrates the generation of diverse solutions and the subsequent critical evaluation. At the core of this framework lies our primary algorithmic contribution, \textbf{Conditional Listwise Policy Optimization (CLPO)}, a reinforcement learning algorithm specifically designed to train the central decision-making policy. This methodology systematically addresses the challenges of credit assignment and signal poverty inherent in complex, language-based collaborative tasks.

\subsection{Preliminaries}
\label{sec:formulation}

We study a round-based collaborative protocol for answering question \(q\). In round \(t\), each of the \(N\) execution agents independently samples $K$ candidates conditioned on the current context
${s_t^{(i)} := (q, b_{t-1}, z_{t-1}^{(i)})}$ for $i \in [N]$,
where
$b_{t-1}$ denotes the previous public
broadcast, and $z_{t-1}^{(i)}$ denotes
the $i$-th agent's private history (state).
%
This yields a total slate \(\mathcal{C}_t\) of \(N\times K\) candidate responses.
The central policy then performs convergence by selecting one candidate
in $\mathcal{C}_t$ to endorse and issues a public broadcast \(b_t\) that contains the index of the answer and optionally a brief justification. The broadcast $b_t$ then conditions the next round $t+1$. This process stops after a fixed number of rounds \(R\),
or when a stopping rule is met. Supervision is primarily the correctness of the endorsed answer at termination, and an optional term rewards the comparative quality of the justification. 
%
%

\subsection{The \Maestro{} Framework: A Paradigm for Collective Synthesis}
\label{sec:MAESTRO}

We now formally introduce the {Multi-Agent Exploration–Synthesis (\Maestro{})} framework.
\Maestro{} operationalizes the divergent–convergent model of creative problem-solving in a principled manner, by decomposing each round of the collaborative process into two distinct phases, which we view through the lenses of Bayesian inference and information theory.

\noindent\textbf{Phase 1: Divergence as Collective Exploration.}
The primary objective of the divergence phase is to effectively explore the vast solution space, mirroring the divergent thinking process. This is achieved through a collective of $N$ parallel Execution Agents. Conditioned on the current state $s_t^{(i)}$, each agent is tasked with generating a diverse set of $K$ candidate solutions. Note here that each candidate solution is a complete trajectory, e.g., a full reasoning chain leading to a final answer. Formally, each agent $i \in [N]$ at
time $t$ samples from its policy $\pi_{\phi^{(i)}}$ as follows:
\begin{align}
    c_{t,k}^{(i)} \;\sim\;\pi_{\phi^{(i)}}(\cdot \mid q,z_{t-1}^{(i)},b_{t-1}), \quad k \in [K].
\label{eq:diverge-sampling}
\end{align}
This collective effort produces candidate pool $\mathcal{C}_t=\{ \{c_{t,k}^{(i)}\}_{k=1}^{K} \}_{i=1}^{N}$. 
%
%
%
%
A key metric for this phase is the \emph{coverage probability ($p_t$)} that the pool contains at least one correct solution:
\begin{equation}
p_t\;:=\;\Pr\!\Big(\bigcup_{i=1}^{N} \bigcup_{k=1}^{K} E(c^{(i)}_{t,k}) \,\Big|\, s_t^{(1)}, \dots, s_t^{(N)} \Big),
\label{eq:coverage}
\end{equation}
where $E(c)$ is the event that candidate $c$ is correct. 
The primary goal of Phase 1 is to increase the expected coverage \(p_t\) in a fixed resource budget. 

\textit{Epsilon-greedy exploration.} To prevent over-conditioning during candidate generation, we allocate a small broadcast-agnostic exploration mass using a simple \emph{epsilon-greedy} strategy.
Specifically, we sample
$\tilde{\pi}_{\phi^{(i)}}(\cdot \mid q, z_{t-1}^{(i)}, b_t) = (1-\varepsilon)\,\pi_{\phi^{(i)}}(\cdot\mid q, z_{t-1}^{(i)},b_{t-1})+\varepsilon\,\pi_{\phi^{(i)}}^{\text{base}}(\cdot\mid q)$ with default $\varepsilon=0.1$,
where $\pi_{\phi^{(i)}}$ is defined in \eqref{eq:diverge-sampling}.
This yields a {coverage floor}: for any subset $A$ of the candidate space, 
$\tilde\pi_{\phi^{(i)}}(A \mid s_t^{(i)}) \ge \varepsilon\,\pi_{\phi^{(i)}}^{\text{base}}(A \mid q)$, so regions reachable by the base policy retain non-zero sampling mass. In practice, we implement the mixture via per-sample random dropout, using the base prompt with probability $\varepsilon$ and otherwise conditioning on the broadcast and the agent’s private history. 

\noindent\textbf{Phase 2: Convergence as List-wise Bayesian Synthesis.}
Following divergent exploration in Phase 1, the convergence phase is orchestrated by a single Central Agent. Its role is to evaluate and synthesize the collective information in the slate \(\mathcal{C}_t\). We view this step as approximating a Bayesian decision over
the posterior probabilities for round $t$:
\begin{equation}
\eta_{t,k}^{(i)} \;:=\; \Pr\big( E(c_{t,k}^{(i)}) \mid q,\mathcal{C}_t\big), \quad i \in [N],\,\, k \in [K].
\label{eq:bayes-argmax}
\end{equation}
Recall that under a \(0\text{--}1\) loss, the Bayes optimal action selects $(i^\star, k^\star) \in \arg\max_{i,k} \eta_{t,k}^{(i)}$. 
We therefore train our Central Agent's policy, $\pi_\theta( \cdot \mid q, \mathcal{C}_t)$, 
to approximate this optimal Bayes decision rule via the CLPO loss
(\Cref{sec:clpo}).
The success of this phase is measured by the \emph{identification probability ($q_t$)}, defined as the conditional probability that the policy selects a correct candidate given that the slate $\mathcal{C}_t$ contains at least one correct option. Specifically, let $S_t := \{ (i, k) \in [N] \times [K] \mid E(c_{t,k}^{(i)}) \textrm{ holds}  \}$ be the latent set of correct candidates and let $(i_t, k_t) \sim \pi_\theta( \cdot \mid q,\mathcal{C}_t)$ denote the centralized decision. Then the identification probability $q_t$ is defined as:
\begin{equation}
q_t \;:=\; \Pr\!\big( (i_t, k_t) \in S_t \,\big|\, q,\mathcal{C}_t,\{|S_t|\ge 1\}\big).
\label{eq:identification}
\end{equation}
This metric quantifies the agent's critical evaluation and synthesis capability.

\noindent\textbf{Broadcast as Public Conditioning.}
After selection in Phase 2, the Central Agent emits a public broadcast $b_t$, containing the endorsed index and a compact justification. 
This broadcast $b_t$ conditions the next round $t+1$.
We can interpret the flow of information as reducing the
Shannon entropy of the ground-truth answer $Y$ with respect
to an observer's posterior;
by the chain rule for mutual information, 
%
we have $H(Y \mid q, b_{1:t}) \leq H(Y \mid q, b_{1:t-1})$ for all $t$, where $b_{1:t} = (b_1, \dots, b_t)$.




\noindent\textbf{Overall Dynamics.}
We summarize the per-round behavior as a coverage–identification factorization conditioned on the public context \((q,b_{t-1})\). The system first attains \emph{coverage} \(p_t\) when the slate contains at least one correct candidate, then achieves \emph{identification} \(q_t\) when the central policy selects a correct candidate.
In Appendix~\ref{app:derivative_cumulative_reliability}, we show the following
cumulative reliability inequality:
if we have that both $p_t \geq \underline{p}$ and $q_t \geq \underline{q}$ almost surely
for all $t$, then $\Pr(\text{success within $R$ rounds}) \geq 1- (1 - \underline{p}\underline{q})^R$.
An immediate consequence is the following tail inequality: if
 $R \geq \frac{1}{\underline{p} \underline{q}} \log\left(\frac{1}{\delta}\right)$,
then the probability of success within the first $R$ rounds is at least $1-\delta$.



\subsection{Conditional Listwise Policy Optimization (CLPO)}
\label{sec:clpo}

Having established the \Maestro{} paradigm, the key question becomes how to \emph{optimize} the convergence process so that the Central Agent can reliably approximate the Bayesian decision rule.\footnote{We do not consider optimizing the policies of the execution agents, which may further improve \Maestro{}.} Conceptually, the two phases of \Maestro{} naturally align with the classical exploration–exploitation trade-off: Phase~1 (divergence) expands the hypothesis space through exploration, while Phase~2 (convergence) serves as exploitation, transforming the diverse candidate set into a single endorsed solution with a supporting rationale. This perspective makes the Phase 2 convergence step a natural fit for reinforcement learning (RL) on the objective:
\begin{equation}
\max_{\pi_\theta}\;\; \mathbb{E}_{(q,\mathcal{C})}\Big[\,r\big(q,\mathcal{C},\text{Chosen},\text{Reason}\big)\,\Big],
\label{eq:clpo-rl-anchor}
\end{equation}
where $\pi_\theta$ denotes the policy of the Central Agent, $\mathcal{C}$ is the candidate pool of responses sub-sampled over all $R$ rounds, and $r$ is our unified reward, which comprises answer correctness and rationale quality assessed via reasoning attributes. This formulation highlights the two challenge at the heart of convergence: the Central Agent must both \textit{(i)} provide a coherent and discriminative rationale that distinguishes the endorsed solution from its competitors, and also \textit{(ii)} select the correct decision token.

\textit{Limitations of Na{\"{i}}ve Sequence-Level Optimization.}
A natural baseline for training \eqref{eq:clpo-rl-anchor}
is \emph{Group Relative Policy Optimization (GRPO)}~\citep{shao2024deepseekmath}, which contrasts candidates within a group and scales sequence log-probabilities by relative advantage. Although suitable for “pick-one-from-$K$” settings, applying GRPO to full sequences exposes three core issues. First, it behaves like \emph{pointwise supervision}: updates treat each completion in isolation rather than judging rationales by their strength {relative} to alternatives. Second, the reward signal is \emph{entangled} across decision and rationale tokens, which obscures credit assignment.
Third, this entanglement induces \emph{spurious style effects}, where verbosity or lexical patterns receive undue credit and concise reasoning is penalized. 

\noindent\textbf{Conditional Listwise Policy Optimization (CLPO).}
We propose {CLPO}, a decoupled training loss that {first} learns to produce reliable, discriminative rationales and {then} learns to make a reliable discrete choice. Concretely, CLPO optimizes the rationale span with a conditional listwise ranking objective~\citep{xia2008listwise} over the entire candidate set.
CLPO allocates the reinforcement signal to the decision tokens, using a focused policy-gradient update to sharpen identification without interference from explanation style. By matching each subproblem to the right objective, CLPO resolves credit entanglement, reduces confounding factors from length/style, and stabilizes training. 
%

\emph{Strategic Decision Loss ($\mathcal{L}_{\text{choice}}$).}
The convergence phase ultimately hinges on the central agent’s ability to make a precise strategic decision: which candidate to endorse. To ensure a clean credit signal, we allocate the reinforcement gradient exclusively to the decision tokens (the choice and its corresponding answer), conditioned on the rationale context. This disentanglement prevents reasoning length or style from interfering with the discrete choice. Formally, we define:
\begin{equation}
    \mathcal{L}_{\text{choice}} = - \mathbb{E}\Big[ \sum_{k=1}^{|\mathcal{C}|} A_k \cdot \log \pi_\theta( k \mid q, \mathcal{C}) \Big], \label{eq:CLPO_choice}
\end{equation}
where the advantage $A_k = r(c_k) - \bar{r}$, and $\bar{r}$ is the average reward within the candidate set. 
In practice, we mask the rationale tokens and aggregate log-probabilities only over the decision segment.

\emph{Tactical Argumentation Loss ($\mathcal{L}_{\text{reason\_rank}}$).}
Agents articulate a justification along with a discrete choice.
In our framework, the rationale is generated \emph{before} the final endorsement. We posit that a justification should not only be plausible on its own, but its plausibility should surpass alternatives. To capture this comparative quality, we employ a \emph{Listwise Ranking Loss}~\citep{xia2008listwise}.
%
Formally, let \(\sigma\) be the permutation that sorts the rewards in descending order, \(r(c_{\sigma_1}) \ge \cdots \ge r(c_{\sigma_{|\mathcal{C}|}})\).
Write the justification for the \(k\)-th candidate in $\mathcal{C}$ 
as a token sequence \(y_{k,1:L_k}\).
We define this loss as:
\begin{align}
    \mathcal{L}_{\text{reason\_rank}} = - \sum_{j=1}^{|\mathcal{C}|} \log \frac{ \exp(s_{\sigma_j}) }{ \sum_{l=j}^{|\mathcal{C}|} \exp(s_{\sigma_l}) }, \quad s_k \;=\; \frac{1}{L_k}\sum_{\tau=1}^{L_k} \log \pi_\theta\!\big(y_{k,\tau}\ \big|\ y_{k,1:\tau-1},\ q,\ \mathcal{C}\big). \label{eq:CLPO_rank}
\end{align}


\noindent\textbf{The CLPO Objective.}
The CLPO training objective combines the two losses 
\eqref{eq:CLPO_choice} and \eqref{eq:CLPO_rank},
with standard regularization terms to ensure stable exploration and prevent catastrophic forgetting. The policy is regularized towards a reference policy $\pi_{\text{ref}}$ (e.g., the initial SFT model) via a KL-divergence term
$\mathcal{L}_{\text{KL}} = \mathbb{E} \left[ D_{\mathrm{KL}}(\pi_\theta(\cdot \mid q, \mathcal{C}) \,\|\, \pi_{\text{ref}}(\cdot \mid q, \mathcal{C})) \right]$
and an entropy bonus $\mathcal{L}_{\text{Entropy}} = \mathbb{E} \left[ H(\pi_\theta(\cdot \mid q, \mathcal{C})) \right]$
that encourages exploration of justifications.
The final objective is:
\begin{equation}
\mathcal{L}_{\text{CLPO}} = \mathcal{L}_{\text{choice}} + \lambda_{\text{rank}} \cdot \mathcal{L}_{\text{reason\_rank}} + \lambda_{\text{kl}} \cdot \mathcal{L}_{\text{KL}} - \lambda_{\text{ent}} \cdot \mathcal{L}_{\text{Entropy}}.
\end{equation}
As we will see shortly, by decoupling the learning objectives for strategic choice and tactical argumentation, CLPO delivers a richer and more stable gradient signal, ensuring clean credit assignment.



\section{Experiments}

\begin{table*}[t]
\small
\centering
\begin{tabular}{lllrrrrrr}
\toprule
\textbf{Type} & \textbf{Mech} & \textbf{Model} & \textbf{GSM8K} & \textbf{MATH} & \textbf{AIME} & \textbf{AMC} & \textbf{MMLU} & \textbf{HumanEval} \\
\midrule
SA  & Ref & Vanilla     & 0.7276 & 0.4285 & 0.0296 & 0.0803 & 0.5799 & 0.4756 \\
SA  & Ref & CoT         & 0.7422 & 0.4693 & 0.0370 & 0.1165 & 0.6157 & 0.5142 \\
SA  & Ref & SC          & 0.8079 & 0.5128 & 0.0407 & 0.1245 & 0.6830 & 0.5752 \\
\midrule
MA  & Prog     & PHP   & 0.8001 & 0.5371 & 0.0444 & 0.1566 & 0.6846 & 0.5650 \\
MA  & Deb     & LLM-Debate  & 0.8352 & 0.5625 & 0.0556 & 0.1928 & 0.6759 & 0.5772 \\
MA  & Deb     & Group-Debate& 0.8398 & \underline{0.5742} & 0.0519 & \underline{0.2048} & \underline{0.6989} & 0.5793 \\
MA  & Dyn     & DyLAN     & 0.8203 & 0.5532 & 0.0370 & 0.1968 & 0.6685 & 0.6159 \\
WF  & Dyn    & GPTSwarm    & \underline{0.8489} & 0.5669 & \underline{0.0578} & 0.1566 & 0.6967 & 0.5955 \\
WF  & Dyn    & AgentPrune  & 0.8438 & 0.5437 & 0.0481 & 0.1647 & 0.6909 & 0.5711 \\
WF  & Dyn      & AFlow    & 0.8375 & 0.5528 & 0.0444 & 0.1205 & 0.6931 & \underline{0.6220} \\
\midrule
WF  & E-S     & \Maestro{}        & 0.8703 & 0.5916 & 0.0556 & 0.2371 & 0.7052 & 0.6267 \\
WF  & E-S     & w/ SFT         & 0.8769 & 0.5983 & 0.0538 & 0.2482 & 0.7085 & 0.6321 \\
WF  & E-S     & w/ GRPO        & 0.8867 & 0.6129 & 0.0704 & 0.2630 & 0.7168 & 0.6538 \\
WF  & E-S     & w/ CLPO        & \textbf{0.8933} & \textbf{0.6285} & \textbf{0.0851} & \textbf{0.2852} & \textbf{0.7238} & \textbf{0.6687} \\
\bottomrule
\end{tabular}
\caption{Comparison of baseline and proposed methods using the LLaMA-8B backbone. The table organizes models by Type (SA: single-agent, MA: multi-agent, WF: workflow-style framework) and by Mechanism (Reflection, Progressive Prompting, Debate, Dynamic Coordination, and Exploration–Synthesis). Underlined numbers indicate the best-performing baseline on each benchmark.
}
\label{tab:main_result}
\end{table*}

\noindent\textbf{Experimental Setup.}
We evaluate our approach across diverse benchmarks, including mathematical reasoning (GSM8K, MATH, AIME, AMC), factual and analytical reasoning (MMLU), and program synthesis (HumanEval), using Solve Rate, Accuracy, and Pass@1 as evaluation metrics. Baselines span single-agent reasoning methods, peer-interaction frameworks, routing and topology controllers, workflow and graph search approaches, and communication-efficient systems. Unless otherwise noted, experiments use three agents and three communication rounds, with instruction-tuned {LLaMA-3B/8B} and {Qwen-3B/7B} models under standard nucleus sampling. All reported results are averaged over three random seeds. See Appendix~\ref{ref:experimental_settings} for a full account of settings.

\subsection{Main Experiments}

\noindent\textbf{Overall Performance.}
Table~\ref{tab:main_result} shows that \Maestro{} consistently surpasses both single-agent and multi-agent baselines across six reasoning benchmarks. On the trainable backbone LLaMA-8B, \Maestro{} with CLPO achieves state-of-the-art accuracy, reaching 89.33\% on GSM8K and 28.52\% on AMC, which corresponds to average gains of 4\%–8\% over strong baselines such as GPTSwarm, AgentPrune, and Group-Debate. The improvements arise from two complementary effects: parallel exploration increases coverage, while CLPO strengthens the central selector’s ability to identify correct solutions. This dual mechanism is especially beneficial on competition-style math tasks such as AMC and AIME, where incorrect but fluent candidates often mislead majority-voting or self-consistency. Importantly, the gains are not limited to trainable backbones. As shown in Table~\ref{tab:main_results_gpt_4o_mini}, even with the closed-source GPT-4o-mini under a prompt-only setting, \Maestro{} achieves the best or tied-best results on GSM8K, MMLU, and HumanEval. The consistency across open- and closed-source models indicates that improvements stem from the collaborative orchestration itself rather than parameter updates, establishing \Maestro{} as a robust paradigm for multi-agent LLM collaboration. 

\begin{table*}[t]
\small
\centering
\begin{tabular}{lrrrrrrrr}
\toprule
\textbf{Dataset} & \textbf{Vanilla} & \textbf{CoT} & \textbf{SC} & \textbf{Debate} & \textbf{GPTS} & \textbf{AP} & \textbf{AF} & \textbf{\Maestro{}} \\
\midrule
GSM8K & 93.17 & 93.68 & 93.32 & 94.66 & 94.66 & \underline{94.89} & 92.30 & \textbf{95.60} \\
MMLU  & 77.81 & 78.43 & 81.05 & 81.04 & 82.80 & 83.02 & \underline{83.10} & \textbf{84.09} \\
HumanEval & 85.71 & 86.69 & 87.58 & 84.38 & 86.28 & 86.80 & \underline{90.06} & \textbf{90.65} \\
\bottomrule
\end{tabular}
\caption{Performance comparison on GSM8K, MMLU, and HumanEval using a {GPT-4o-mini} backbone. 
{\Maestro{}} consistently achieves the highest accuracy across all benchmarks, outperforming both single-agent methods (Vanilla, CoT, and SC) and existing multi-agent frameworks like Debate, GPTSwarm (GPTS), AgentPrune (AP), and AFlow (AF). The improvements confirm that \Maestro\ remains effective even when applied zero-shot to closed-source LLMs.
}
\label{tab:main_results_gpt_4o_mini}
\end{table*}

\noindent\textbf{Cross-Backbone Consistency.}
To further examine the generality of our optimization strategy, we applied \Maestro{} with CLPO across different LLM backbones, including LLaMA-8B, LLaMA-3B, Qwen-7B, and Qwen-3B (Table~\ref{tab:baseline_backbone}).
We observe consistent improvements across all settings. On GSM8K, the accuracy reaches 89.33\% with LLaMA-8B, 81.53\% with LLaMA-3B, 95.12\% with Qwen-7B, and 90.83\% with Qwen-3B, establishing clear gains compared to their strongest respective baselines. The effectiveness of CLPO is not confined to a specific model family or size. Instead, as shown in Table~\ref{tab:llm_backbone_acc_coverage}, the optimization consistently enhances the identification probability $q_t$, enabling the central synthesis agent to more reliably distinguish correct solutions from plausible distractors. Importantly, this pattern is also reflected on AMC, where the improvements are similarly pronounced, underscoring that the collaborative mechanism combined with CLPO is broadly transferable across architectures. Overall, these findings confirm that \Maestro{} with CLPO achieves robust gains across backbones, validating the universality of our collaborative optimization paradigm.

\begin{table*}[t]
\small
\centering
\begin{tabular}{lrrrr}
\toprule
\textbf{Model} 
 & \textbf{LLaMA-8B} 
 & \textbf{LLaMA-3B} 
 & \textbf{Qwen-7B} 
 & \textbf{Qwen-3B} \\
\midrule
Vanilla    & 0.7276 & 0.4685 & 0.9088 & 0.8337 \\
CoT        & 0.7422 & 0.5014 & 0.9098 & 0.8456 \\
SC         & 0.8079 & 0.5421 & 0.9295 & 0.8860 \\
\midrule
PHP        & 0.8001 & 0.6222 & 0.9330 & 0.8645 \\
LLM-Debate & 0.8352 & 0.7584 & \underline{0.9363} & 0.8714 \\
DyLAN      & 0.8203 & \underline{0.7647} & 0.9315 & \underline{0.8810} \\
GPTSwarm   & \underline{0.8489} & 0.6919 & 0.9227 & 0.8678 \\
AgentPrune & 0.8438 & 0.6502 & 0.9244 & 0.8643 \\
AFlow      & 0.8375 & 0.6837 & 0.9286 & 0.8752 \\
\midrule
\Maestro{} w/ CLPO & \textbf{0.8933} & \textbf{0.8153} & \textbf{0.9512} & \textbf{0.9083} \\
\bottomrule
\end{tabular}
\caption{Performance of collaborative reasoning baselines across four backbone LLMs (LLaMA-8B, LLaMA-3B, Qwen-7B, Qwen-3B) on GSM8K. \Maestro{} w/ CLPO consistently achieves the highest accuracy, demonstrating robustness and generality across model architectures.}
\label{tab:baseline_backbone}
\end{table*}

\subsection{Analysis Experiments}

\begin{figure}[t]
  \centering
  \begin{tabular}{cc}
    \includegraphics[width=.48\linewidth]{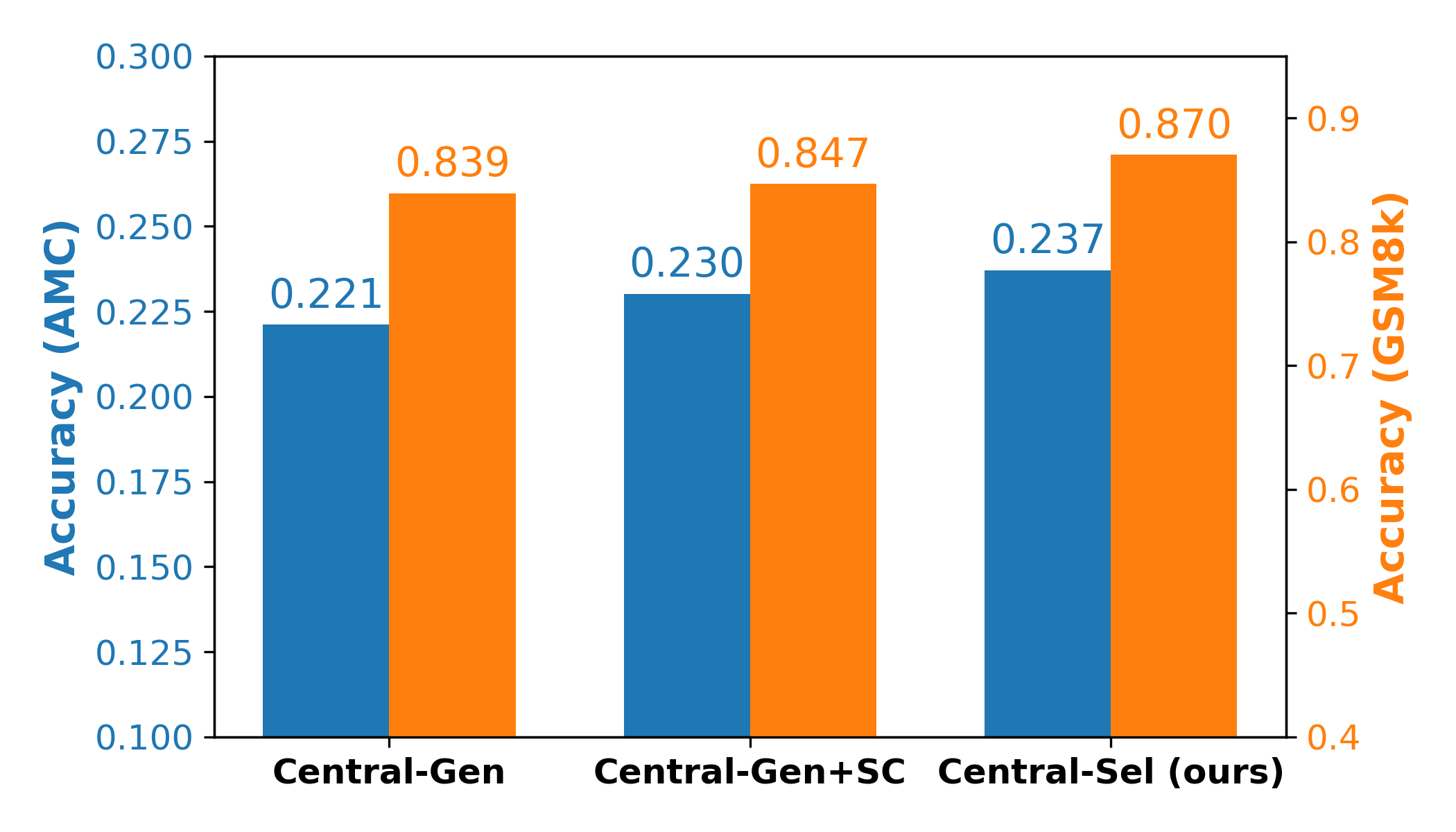}
    &
    \includegraphics[width=.48\linewidth]{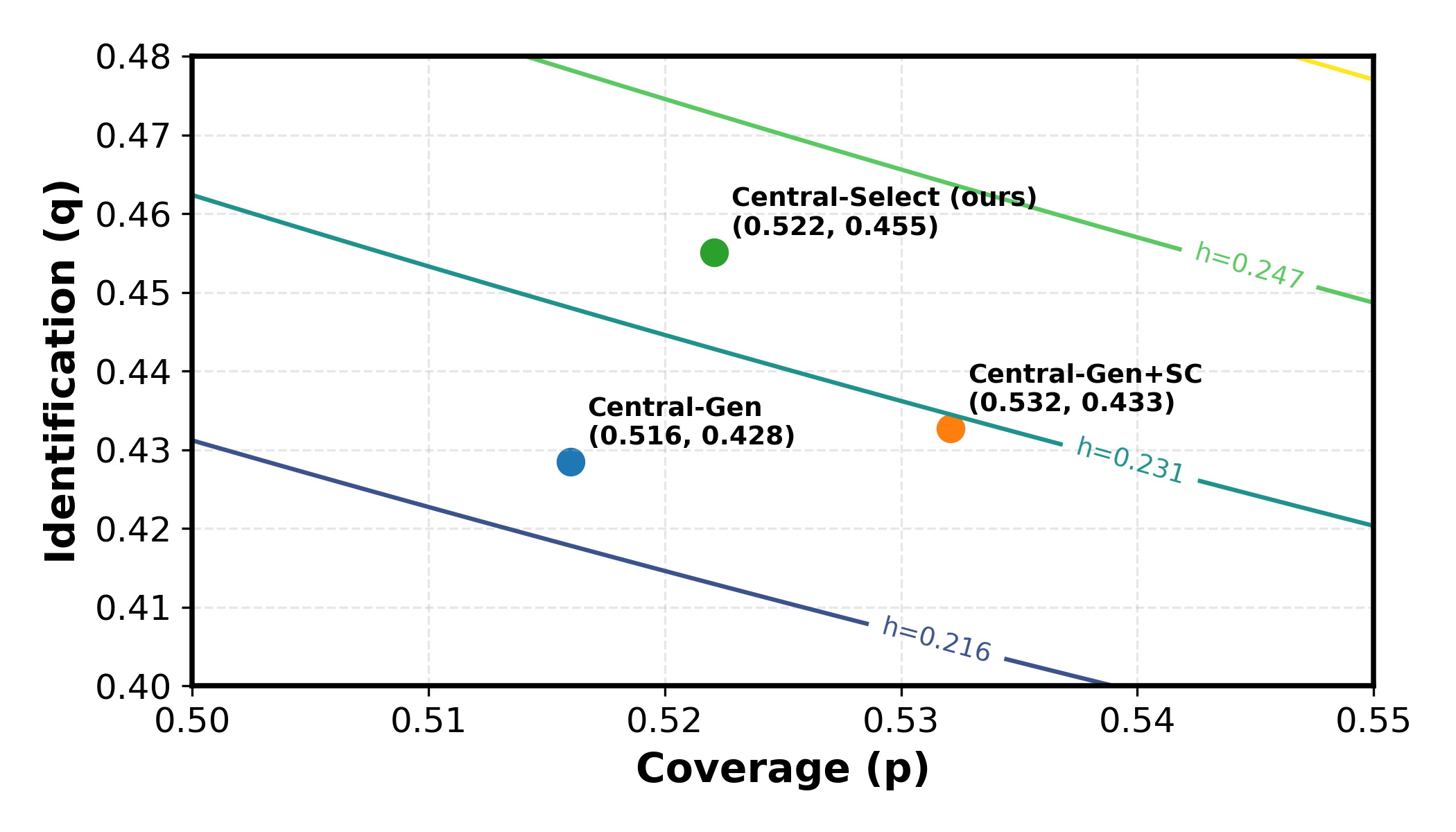}
  \end{tabular}
  \vspace{-12pt}
  \caption{Comparison of collaboration paradigms. \textbf{Left:} task accuracy on AMC and GSM8K across different central coordination strategies. \textbf{Right:} performance decomposed into coverage and identification rates; central selection transforms diverse reasoning into reliable outcomes.}\label{fig:variant_paradigm_performance}
\end{figure}

\noindent\textbf{Centralized Paradigm Variants: Selection vs. Generation.}
We now examine two natural centralized paradigms for convergence, namely generation and selection, to clarify why the latter forms the core of \Maestro{} (Figure~\ref{fig:variant_paradigm_performance}).
When the central agent directly generates a reasoning trajectory and final answer (\textsc{Central-Gen}), the accuracy drops substantially. Incorporating self-consistency into generation (\textsc{Central-Gen+SC}) yields only a marginal gain. In contrast, the selection paradigm (\textsc{Central-Select}, ours) described in \Cref{sec:MAESTRO} attains the highest accuracy, reaching 0.870 on GSM8K and 0.237 on AMC. Figure~\ref{fig:variant_paradigm_performance} (Right) further decomposes the results into coverage and identification probabilities. The three paradigms exhibit similar coverage, but the identification probability is markedly higher under \textsc{Central-Select} (0.8919 on GSM8K and 0.4551 on AMC), which directly explains its superior end-to-end accuracy. We hypothesize that generation forces the central agent to absorb long and noisy contexts from multiple candidates, often diluting critical distinctions and amplifying misleading patterns. LLMs also tend to prioritize narrative coherence over factual correctness, which makes direct generation vulnerable to self-consistent hallucinations~\citep{farquhar2024detecting,banerjee2025llms}. Self-consistency mitigates randomness but cannot overcome these structural issues. By contrast, the selection paradigm frames convergence as a discriminative comparison among competing candidates, thereby preserving informative differences and reliably elevating correct solutions. 
Figure~\ref{fig:ab_sankey} (appendix) contains a complementary visualization. 

\begin{figure}[t]
  \centering
  \begin{tabular}{cc}
    \includegraphics[width=.48\linewidth]{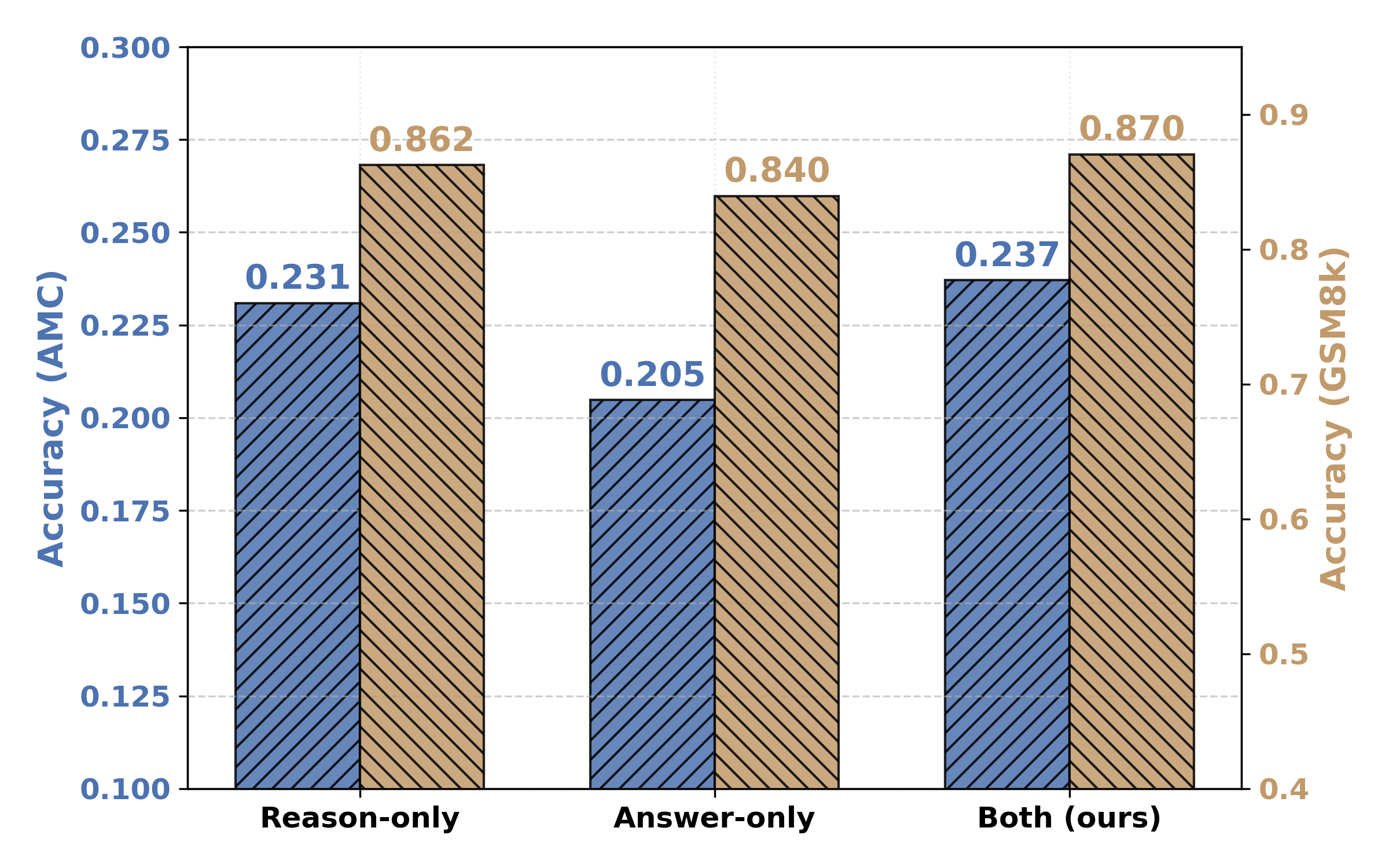}
    &
    \includegraphics[width=.48\linewidth]{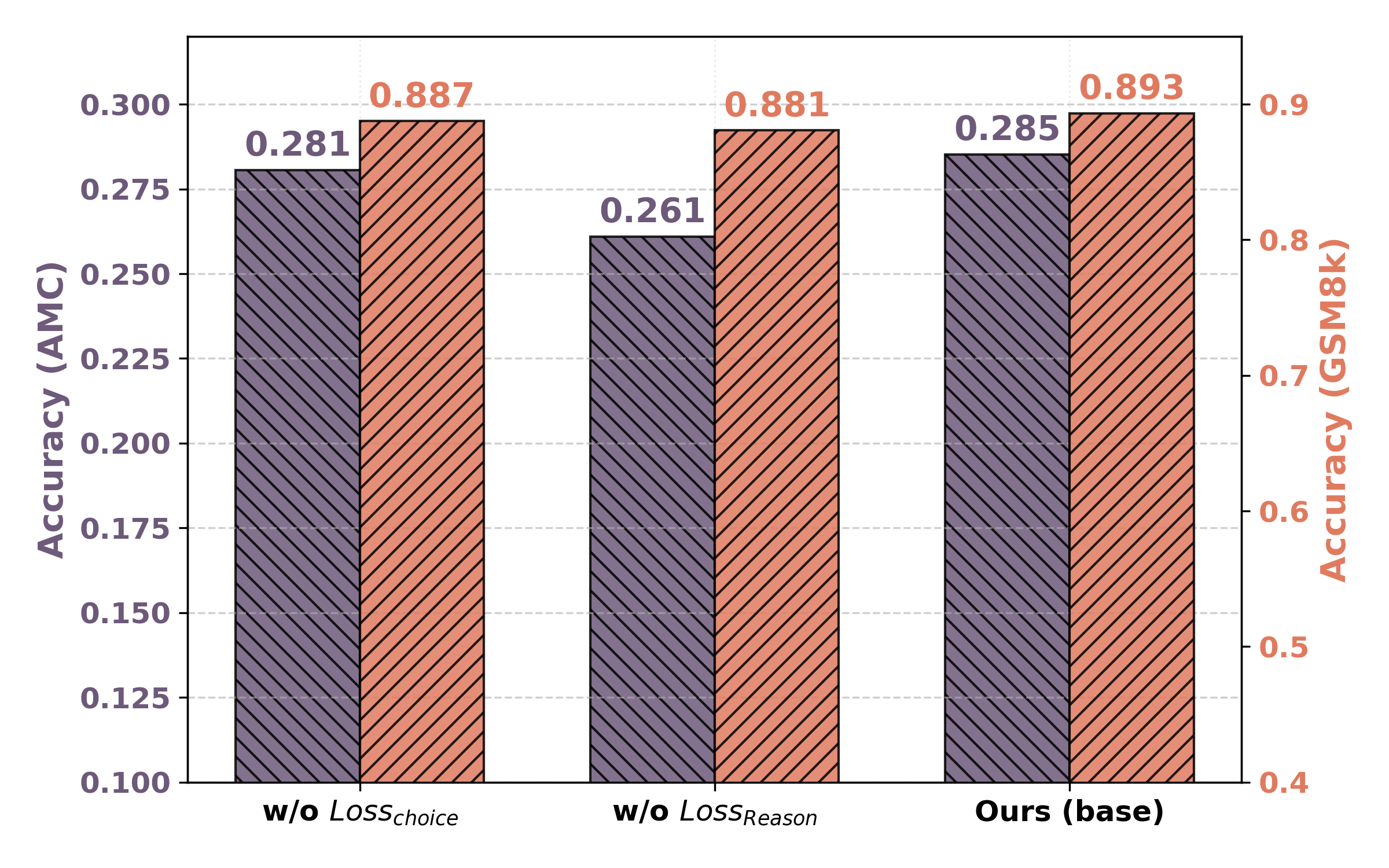}
  \end{tabular}
  \vspace{-12pt}
  \caption{Ablation studies on central selection inputs and CLPO losses. \textbf{Left:} Reason-only, Answer-only, and Both settings when passing candidate information to the central selector. 
  \textbf{Right:} contributions of loss components studied by removing choice or reasoning supervision.}
  \label{fig:reason_answer_tradeoff}
\end{figure}

\noindent\textbf{Evidence versus Verdict in Centralized Selection.}
We examine how different types of candidate information influence the central selector’s decisions by comparing three settings: \emph{Reason-only} (only reasoning steps), \emph{Answer-only} (only the final answer), and \emph{Both} (ours, reasoning with the answer). As shown in Figure~\ref{fig:reason_answer_tradeoff}, \emph{Answer-only} yields the weakest performance (GSM8K $0.840$, AMC $0.205$), while \emph{Reason-only} performs better but remains slightly below the full setting. Since the candidate pool is identical, coverage is unchanged and differences arise from identification capability. The results show that reasoning and outcomes play complementary roles. Without reasoning, the selector lacks evidential structure and often falls back on superficial heuristics. Without final outcomes, it struggles to resolve cases where plausible reasoning paths diverge to different answers. Combining both provides the strongest performance: reasoning paths supply discriminative evidence, while answers anchor the verdict and disambiguate close cases. 

\noindent\textbf{Disentangled Optimization Signals in CLPO.}
To better understand the contribution of each optimization signal in CLPO, we conduct an ablation study by removing either the decision-focused loss $\mathcal{L}_{\text{choice}}$ or the rationale ranking loss $\mathcal{L}_{\text{reason}}$ (Figure~\ref{fig:reason_answer_tradeoff}).
Removing $\mathcal{L}_{\text{choice}}$ causes a modest decline, indicating that ranking-based supervision over rationales alone can sustain reasonable convergence. In contrast, removing $\mathcal{L}_{\text{reason}}$ leads to a sharp degradation (AMC $0.261$, GSM8K $0.881$); without comparative evaluation of explanations, the selector is more easily swayed by persuasive but incorrect candidates. The full CLPO objective achieves the best performance, confirming the necessity of combining both terms. This pattern aligns with our design intuition: $\mathcal{L}_{\text{choice}}$ strengthens decisiveness by refining the probability of endorsing the correct candidate, while $\mathcal{L}_{\text{reason}}$ enforces discriminative evidence quality by forcing correct rationales to outrank distractors. Their joint effect provides clean credit assignment across decisions and justifications, ensuring that convergence is accurate.

\subsection{Hyper-parameter Analysis}

\begin{figure}[t]
  \centering
  \begin{tabular}{cc}
    \includegraphics[width=.48\linewidth]{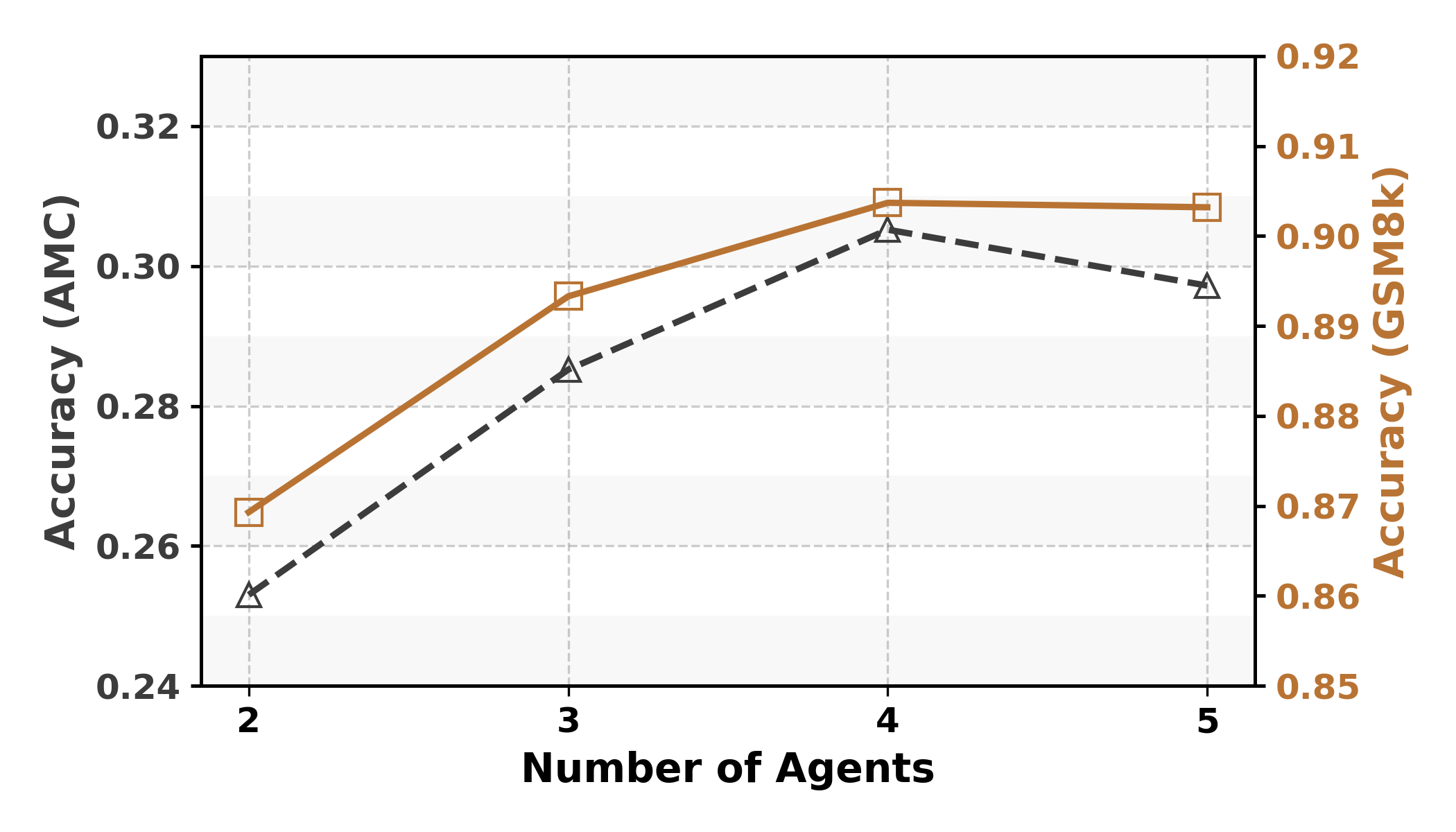}
    &
    \includegraphics[width=.48\linewidth]{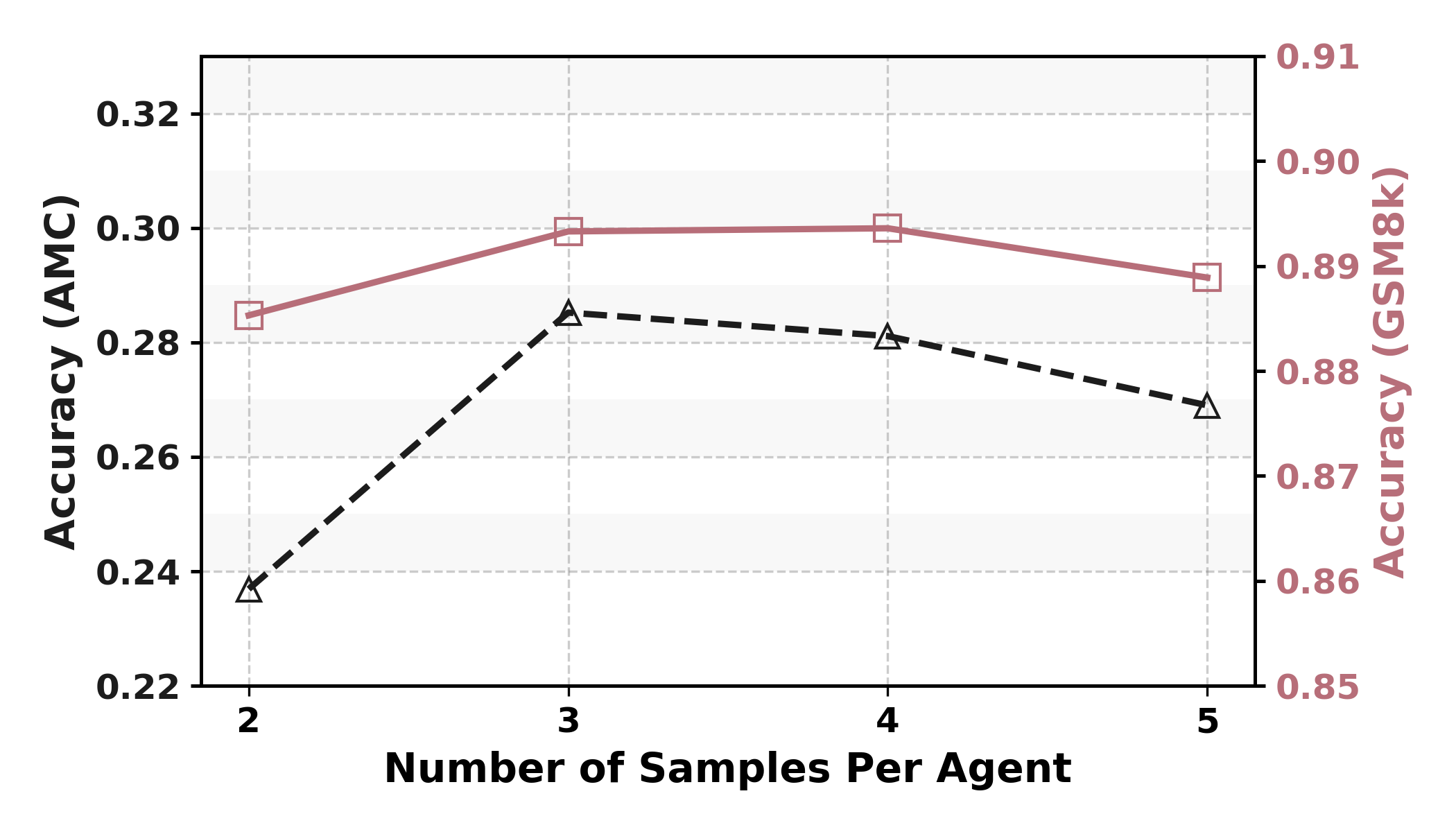}
  \end{tabular}
  \vspace{-12pt}
  \caption{Effect of collaborative scale on reasoning performance. 
  \textbf{Left:} accuracy with varying agent numbers.
  \textbf{Right:} impact of sampling multiplicity. 
  In each, the solid line corresponds to GSM8K accuracy (right axis) and the dashed line corresponds to AMC accuracy (left axis).
  }
  \label{fig:number_agents_k}
\end{figure}

\noindent\textbf{Scaling Agent Populations and Collaboration Rounds.}
We study how the size of the agent collective and the number of collaboration rounds influence performance. As shown in Figure~\ref{fig:number_agents_k}, increasing the population from two to four steadily improves accuracy (AMC $0.253 \to 0.3052$; GSM8K $0.8693 \to 0.9037$). Broader exploration raises the probability that at least one candidate is correct, and the central selector trained with CLPO can convert this coverage into higher identification accuracy. Beyond four agents, however, gains saturate and slightly decline because redundancy introduces distractors. A similar trend appears when varying the number of collaboration rounds (see Figure~\ref{fig:ab_rounds} in the appendix). Adding one or two rounds improves identification by focusing exploration through broadcasted evidence, while coverage changes little once the initial pool is large. Excessive rounds reduce diversity, amplify early errors through herding, and increase stochastic variance. The results highlight a consistent trade-off: additional agents and rounds enhance coverage and identification up to a point, but beyond that redundancy and bias dominate. Moderate settings of 3--4 agents and 2--3 rounds achieve the best balance. A more detailed analysis is provided in Appendix~\ref{app:exp_ab_agent_round}.

\noindent\textbf{Sampling Depth per Agent: Coverage–Variance Trade-off.}
We examine how the number of samples per agent (\(K\)) affects performance (Figure~\ref{fig:number_agents_k}). 
As \(K\) increases from 2 to 3, accuracy improves (AMC \(0.2369 \to 0.2852\); GSM8K \(0.8853 \to 0.8933\)). After, gains saturate: GSM8K changes little at \(K{=}4\) and declines at \(K{=}5\), while AMC peaks at \(K{=}3\) before dropping, reflecting the exploration–synthesis decomposition. Larger \(K\) initially raises coverage, but multi-sampling from the same policy quickly yields correlated and redundant outputs;
deeper sampling inflates within-round variance by drawing repeatedly from one agent rather than diversifying across agents. Once coverage nears saturation, additional samples contribute more noise than signal. These results suggest a practical guideline: allocate budget to enlarging the number of agents to diversify hypotheses, while keeping \(K\) modest so that the selector can reliably convert coverage into identification. 
A more detailed analysis is provided in Appendix~\ref{app:exp_ab_sample_size}.

\section{Conclusion}
We introduce \Maestro{}, a principled framework for multi-agent collaboration that enables both divergent exploration and convergent synthesis. We also present CLPO, an RL method that achieves precise credit assignment through decision-focused optimization and comparative supervision. Together these components yield consistent improvements across diverse reasoning benchmarks and surpass state-of-the-art multi-agent methods. Looking ahead, we plan to investigate unified policy objectives that jointly optimize exploration and synthesis, and continuous learning paradigms that enable multi-agent collectives to refine collaboration dynamics through self-improvement over time.

\bibliography{iclr2026_conference}
\bibliographystyle{iclr2026_conference}

\newpage
\appendix

\section{A Comprehensive Review of Related Work}
\label{ref:related_work}

\subsection{Multi-Agent LLM Collaboration}

Single LLM agents, despite their impressive individual capabilities, face fundamental limitations in context length, sequential generation, and breadth of expertise. These constraints hinder performance on tasks that demand parallel information processing, complementary skill sets, and the synthesis of diverse perspectives~\citep{gabriel2024advancing, liang2023encouraging, xiong2023examining, yin2023exchange, zhang2023exploring}. To overcome these bottlenecks, researchers have increasingly turned to multi-agent systems (MAS), where collectives of LLM-powered agents coordinate to realize forms of collective intelligence in domains such as software engineering, complex planning, and scientific discovery~\citep{chang2025survey,ye2024domain, chen2023agentverse, ning2023skeleton, ping2025verimoa,ping2025hdlcore, pan2024agentcoord, suzgun2024meta, chen2023autoagents, ishibashi2024self}.

Early approaches largely follow a \emph{prompt-based paradigm}, where roles, protocols, and workflows are specified by hand. Debate-style and critique frameworks~\citep{du2023improving,chan2023chateval,chen2024llmarena,mukobi2023welfare,wang2023avalon,abdelnabi2024cooperation} as well as corporate-style pipelines such as MetaGPT~\citep{hong2023metagpt,qian2023chatdev} exemplify this direction. These systems demonstrate the promise of structured collaboration but remain brittle because their strategies are statically prescribed and cannot adapt or learn from experience~\citep{jiang2023llm,liang2023encouraging,he2023lego}. 

Beyond static prompting, recent work introduces more principled coordination schemes. \emph{Prestructured paradigms} adopt fixed interaction topologies, such as chains, trees, or graphs, to organize communication and enforce critique~\citep{du2023improving,liu2024groupdebate,qian2024scaling}. In parallel, \emph{self-organizing paradigms} dynamically adapt collaboration graphs during inference using search, pruning, or routing methods, as seen in DyLAN, MasRouter, GPTSwarm, and AFLOW~\citep{liu2023dynamic,hu2024automated,shang2024agentsquare,zhang2024cut,zhuge2024gptswarm,zhang2024aflow,hu2024routerbench,yue2025masrouter}. These frameworks improve efficiency and flexibility, yet they often reduce coordination to architectural wiring and lack mechanisms for fine-grained credit assignment.

Complementary efforts focus on \emph{role specialization} and \emph{organizational analogies}, where agents are differentiated as planners, solvers, or verifiers, or even structured as corporate roles such as CEO and engineer~\citep{hong2023metagpt,li2023camel,mandi2024roco,du2023improving,chen2023agentverse}. Communication protocols vary between centralized, decentralized, and hierarchical settings, as well as synchronous versus asynchronous exchanges~\citep{jiang2023llm,ning2023skeleton,pan2024agentcoord,liang2023encouraging,zhang2023exploring,du2023improving,chan2023chateval,chen2024llmarena}. These design choices trade off scalability, robustness, and overhead but leave unresolved the fundamental question of how to separate decision-making from justification in a principled manner. 

Overall, existing MAS paradigms illuminate diverse strategies for orchestrating collaboration, but they remain limited in their ability to balance broad exploration with reliable convergence and to assign credit cleanly across agents and rationales.

\subsection{Reinforcement Learning for Multi-Agent LLMs.}
A central trend in multi-agent LLM research is to move beyond static prompt engineering toward learning from interaction. Early work explored supervised fine-tuning (SFT) on expert demonstrations, which injects cooperative behaviors by imitation but is limited in adaptability to unseen coordination settings \citep{madaan2023self, zelikman2024quiet}. In contrast, reinforcement learning (RL) supplies a reward-driven mechanism that allows agents to refine strategies from experience and discover emergent collaboration patterns \citep{zhu2025lamarl, zhuang2024yolo}. In practice, SFT often initializes base policies, while multi-agent reinforcement learning (MARL) further tailors them under task feedback \citep{zhu2025lamarl, zhang2021multi}.

Recent efforts fall into three complementary directions. First, some approaches compile language into structured controllers before learning, such as translating dialogue into plans, graphs, or code, which grounds RL optimization in compact symbolic spaces \citep{zhuang2024yolo, jia2025enhancing}. Second, others focus on adaptive collaboration online, dynamically refining task decomposition, agent assignment, or communication routing through RL signals \citep{zhou2025reso, wang2024self, xu2025scalable}. Third, direct policy optimization for reasoning behaviors has gained traction, with GRPO- and PPO-style updates applied to cooperative justification and answer selection, often combined with tool use or human feedback \citep{wan2025rema, park2025maporl, han2025joyagents}. Across these directions, RL provides the flexibility to align multi-agent dynamics with task objectives rather than relying solely on fixed prompts or wiring rules.

At the same time, this line of work highlights several core challenges. A prominent difficulty is credit assignment: linguistic outputs entangle the correctness of discrete decisions with the plausibility of accompanying rationales, making it unclear what aspect of behavior is being rewarded \citep{wei2025lero, jiang2025qllm}. Another challenge is efficient exploration in vast language action spaces, where agents may generate superficially diverse but semantically redundant outputs \citep{liu2023forward, zhang2024multi}. Finally, there is the issue of alignment of emergent behaviors, since collaboration can amplify biases or drift without proper reward shaping \citep{alsadat2024multi, lin2025speaking}.

Our work follows this trajectory while placing a sharper emphasis on the convergence step of collaboration. Rather than treating group outcomes as a monolithic reward signal, we recast convergence as a structured optimization problem that separates the supervision of rationales from decision signals. This perspective motivates the design of a new RL objective that provides comparative supervision across rationales while preserving clean decision gradients, complementing existing GRPO-style multi-agent optimization.

\section{Derivation of the Cumulative Reliability Inequality}
\label{app:derivative_cumulative_reliability}

In this section we provide a derivation of the cumulative reliability inequality
from \Cref{sec:MAESTRO}.

We start by defining the history (i.e., filtration) $\mathcal{F}_t := (q, \theta_{1:t}, \zeta_{1:t})$, with $\mathcal{F}_0 := q$,
where $\theta_{1:t}$ denotes the randomness of the execution agents for the first $t$ rounds,
and $\zeta_{1:t}$ denotes the randomness of the central agent for the first $t$ rounds.
Let \(\mathsf{Cand}_t := \{\exists\, (i, k) \in [N] \times [K] \text{ s.t. $E(c^{(i)}_{t,k})$ holds}\}\) denote the event the round \(t\) slate $\mathcal{C}_t$ contains at least one correct candidate. Let \((i_t, k_t) \sim \pi_\theta(\cdot \mid q,\mathcal{C}_t)\) be the central decision made at time $t$. 
Then we have that, assuming $p_t \geq \underline{p}$
and $q_t \geq \underline{q}$ for non-random $\underline{p},\underline{q}$ almost surely for all $t$:
\begin{align*}
    h_t &:= \Pr(\text{Success}_t \mid \mathcal{F}_{t-1}) \\
    &= \E_{\mathcal{C}_t \mid \mathcal{F}_{t-1}}[ \Pr(\text{Success}_t \mid \mathcal{F}_{t-1}, \mathcal{C}_t )] \\
    &\stackrel{(a)}{=} \E_{\mathcal{C}_t \mid \mathcal{F}_{t-1}}[ \mathbf{1}\{ \mathsf{Cand}_t \} \Pr( (i_t, k_t) \in S_t \mid q, \mathcal{C}_t, \{ |S_t| \geq 1 \}) ] \\
    &\stackrel{(b)}{=} \E_{\mathcal{C}_t \mid \mathcal{F}_{t-1}}[ \mathbf{1}\{ \mathsf{Cand}_t \}  q_t ] \\
    &\stackrel{(c)}{\geq} \E_{\mathcal{C}_t \mid \mathcal{F}_{t-1}}[ \mathbf{1}\{ \mathsf{Cand}_t \} ] \underline{q} \\
    &= \Pr( \mathsf{Cand}_t \mid \mathcal{F}_{t-1}) \underline{q} \\
    &\stackrel{(d)}{=} \Pr( \mathsf{Cand}_t \mid q, s_t^{(1:N)}) \underline{q} \\
    &\stackrel{(e)}{=} p_t \underline{q} \\
    &\stackrel{(f)}{\geq} \underline{p} \underline{q},
\end{align*}
where in (a) we used the fact that the decision $(i_t, k_t)$ is generated conditioned only on $(q, \mathcal{C}_t)$,
(b) is the definition of $q_t$ from \eqref{eq:identification},
(c) uses our lower bound assumption on $q_t$,
(d) uses the fact that the candidate decisions $\mathcal{C}_t$ are generated
conditioned on $(q, z_{t-1}^{(1:N)}, b_{t-1})$, which is contained within $\mathcal{F}_{t-1}$,
(e) is the definition of $p_t$ from \eqref{eq:coverage},
and (f) uses our lower bound assumption on $p_t$.

Now, let $X_t := \mathbf{1}\{\text{Success}_t\}$.
By definition we have that $X_t$ is $\mathcal{F}_t$-measurable.
Hence by repeated applications of the tower-property of conditional expectations,
\begin{align*}
    \Pr( \text{Fail all $R$ rounds} ) &= \mathbb{E}\left[ \prod_{t=1}^{R} (1 - X_t) \right] \\
    &= \mathbb{E}\left[ \mathbb{E}\left[  \prod_{t=1}^{R} (1 - X_t)   \mid \mathcal{F}_{R-1} \right]  \right] \\
    &= \mathbb{E}\left[ \prod_{t=1}^{R-1} (1-X_t) \mathbb{E}\left[ 1 - X_R \mid \mathcal{F}_{R-1} \right]   \right] \\
    &= \mathbb{E}\left[ \prod_{t=1}^{R-1} (1-X_t) (1 - h_R) \right] \\
    &\stackrel{(a)}{\leq} \mathbb{E}\left[ \prod_{t=1}^{R-1} (1-X_t) \right] (1 - \underline{p} \underline{q}) \leq \dots \leq (1 - \underline{p} \underline{q})^R,
\end{align*}
where (a) follows from above where we established $h_t \geq \underline{p} \underline{q}$.

\section{Experiment}

\subsection{Experimental Settings}
\label{ref:experimental_settings}

\paragraph{Datasets \& Benchmarks.}
We evaluate the framework on three task families designed to stress complementary aspects of collective reasoning, namely precise numeric inference, broad factual and analytical judgment, and executable synthesis. This spectrum assesses both the “diverge” capacity (hypothesis coverage) and the “converge” capacity (principled selection).

\emph{Mathematical reasoning.} We use \textbf{GSM8K}, \textbf{MATH}, \textbf{AIME}, and \textbf{AMC}. GSM8K comprises grade-school word problems with single numeric targets; MATH covers competition-level problems across algebra, number theory, geometry, and combinatorics; AIME consists of short-answer olympiad items with integer solutions; AMC includes large-scale contest questions (we report on the standard subset with unambiguous numeric targets). Performance is measured by \emph{Solve Rate}, the proportion of items whose predicted answer exactly matches the ground truth under benchmark normalization rules.

\emph{General reasoning.} We use \textbf{MMLU}, spanning 57 subjects from STEM to humanities under a four-choice multiple-choice format. Performance is reported as \emph{Accuracy}, i.e., the fraction of correctly selected options, under the benchmark’s standard few-shot setting.

\emph{Code generation.} We use \textbf{HumanEval}, where models synthesize functions from natural-language specifications. Performance is reported as \emph{Pass@1}, the percentage of prompts for which the single generated solution passes all hidden unit tests.

Unless otherwise noted, we follow official splits and prompting guidelines, do not use external tools or retrieval, and keep evaluation deterministic for single predictions. When stochastic sampling is required (e.g., for self-consistency or multi-agent generation), we fix seeds and average over repeated runs; confidence intervals are reported in the appendix. This protocol ensures comparability with prior work while isolating the contribution of the collaboration paradigm and training objective.

\paragraph{Baselines.}
We compare against collaborative LLM methods organized by their underlying \emph{collaboration mechanism}, rather than model brand. 
(i) \emph{Single-agent reasoning}: \textbf{Vanilla} (direct decoding), \textbf{CoT} (chain-of-thought prompting), and \textbf{SC} (self-consistency with majority vote). 
(ii) \emph{Peer interaction}: \textbf{LLM-Debate} (multi-round argumentation with shared transcripts), \textbf{GroupDebate} (multi-agent debate with voting-based aggregation) and \textbf{PHP} (pairwise critique without a global selector). 
(iii) \emph{Routing/topology control}: \textbf{DyLAN} (layered agent network with pruning and early-stop consensus). 
(iv) \emph{Workflow/graph search}: \textbf{GPTSwarm} (optimization of reasoning graphs over multiple prompting strategies) and \textbf{AFLOW} (Monte-Carlo search over reusable operators). 
(v) \emph{Communication efficiency}: \textbf{AgentPrune} (sparse message passing to reduce cost while maintaining accuracy). For all baselines we use the same base models, adopt each method’s official prompts and stopping criteria, and match collaboration budgets (rounds, agents, and generations). When methods output multiple candidates, we apply their canonical aggregation (e.g., majority vote or ranker). This taxonomy clarifies whether improvements come from stronger generation (divergence), more reliable selection (convergence), or better workflow, providing a diagnostic comparison to our exploration–synthesis paradigm.

\paragraph{Prompt Templates.}
To make our experimental setup transparent and reproducible, we explicitly document the 
instruction prompts used by different agents in our framework. These templates capture the 
roles and responsibilities of both reasoning agents and the center arbiter, highlighting how 
they collaborate through structured interaction. For clarity, we present concrete examples 
in the domain of mathematical reasoning problems, which serve as a representative 
case for illustrating the prompt design.

\begin{tcolorbox}[title={Execution Agent Prompt (Initial Round)}]
You are Reasoning Agent \#\{agent\_id\}.  
Your task is to carefully solve the given math problem step by step.  
Clearly show your reasoning process, making sure that each transformation is logically valid.  
Avoid skipping important intermediate steps.  

At the end of your reasoning, provide the final numeric answer in the exact format: \verb|\boxed{...}|.  

\textbf{Problem:} \{\{math\_question\}\}
\end{tcolorbox}

\begin{tcolorbox}[title={Execution Agent Prompt (Interactive Round)}]
You are Reasoning Agent \#\{agent\_id\}.  
You previously proposed multiple solutions and now also receive the Center Arbiter's synthesis.  

Re-evaluate the problem carefully, considering both your earlier solutions and the Arbiter's feedback.  
Generate refined solutions that correct any mistakes if needed, ensuring logical consistency.  

Each output must end with the final numeric answer in the exact format: \verb|\boxed{...}|.  

\textbf{Problem:} \{\{math\_question\}\}  \newline
\textbf{Your Previous Solutions:} \{...\}  \newline
\textbf{Center Arbiter's Feedback:} \{...\} \newline
\end{tcolorbox}

\begin{tcolorbox}[title={Central Agent Prompt}]
You are the Center Arbiter, responsible for evaluating candidate solutions proposed by agents.  
Carefully read the original problem and all candidate solutions.  
Compare their reasoning, detect mistakes if present, and identify the most reliable candidate.  

Then, following the strict format below, provide a short justification, the chosen candidate index,  
and the final numeric answer in \verb|\boxed{...}|.  

\textbf{Problem:} \{\{math\_question\}\}  

\textbf{Candidates:} \newline
- Candidate 1: \{...\} \newline
- Candidate 2: \{...\} \newline
- Candidate 3: \{...\} \newline

\textbf{STRICT OUTPUT FORMAT:}  \newline
Reason: \{detailed justification\} \newline
Chosen: \{candidate\_id\} \newline
Final: \verb|\boxed{...}|  

\end{tcolorbox}

\paragraph{Implementation Details.}
Our experiments are conducted with a compact configuration where three agents interact across three communication rounds for each query. The agents are instantiated from widely used instruction-tuned models including \textbf{Llama-3.1-8B-Instruct}, \textbf{Llama-3.2-3B-Instruct}~\citep{dubey2024llama}, and \textbf{Qwen2.5-7B-Instruct} as well as \textbf{Qwen2.5-3B-Instruct}~\citep{team2024qwen2}. All models are accessed through the HuggingFace Transformers library with 8-bit quantization to reduce GPU memory usage. We enable KV caching throughout the experiments to improve generation efficiency. We adopt a unified generation setup across all experiments. Unless otherwise noted, nucleus sampling with $p=0.95$ is used and the maximum output length is set to $512$ tokens. The default temperature is $0.7$, which balances diversity and stability. For tasks requiring deterministic evaluation, such as pairwise preference comparisons or revision prompts, we reduce the temperature to $0.3$. The central agent is always assigned a temperature of $0.0$ to enforce deterministic decisions and avoid stochastic drift. To ensure comparability across methods, all models share the same decoding settings and random seeds are fixed. This setup follows common practice in LLM evaluation and ensures that performance differences stem from the collaboration paradigm rather than decoding hyperparameters. During both supervised fine-tuning (SFT) and policy optimization, we adopt parameter-efficient fine-tuning using LoRA. Unless otherwise noted, the LoRA rank is set to $16$, with scaling factor $\alpha=32$ and dropout $0.05$. Only LoRA parameters, LayerNorm statistics, and bias terms are updated, while all other weights remain frozen. Training uses Adam ($\beta_1=0.9$, $\beta_2=0.999$, $\epsilon=10^{-8}$) with an initial learning rate of $5\times10^{-5}$, decayed following a cosine schedule. Gradient norms are clipped at $1.0$ to stabilize optimization. We train with a global batch size of $32$ distributed across four A100 GPUs (80GB each), using mixed precision (bfloat16) for efficiency. The rank-loss coefficient is tuned over \{0.1, 0.5, 0.8, 1.0\}. To encourage exploration we add an entropy bonus of $0.01$, while maintaining consistency with the SFT reference policy via a KL regularization weight of $0.1$. Each run proceeds for three epochs, and results are averaged over three random seeds (25, 42, and 99) to ensure robustness and mitigate variance.

\subsection{Experimental Results.}

\begin{figure}[h]
    \centering
    \includegraphics[width=0.8\linewidth]{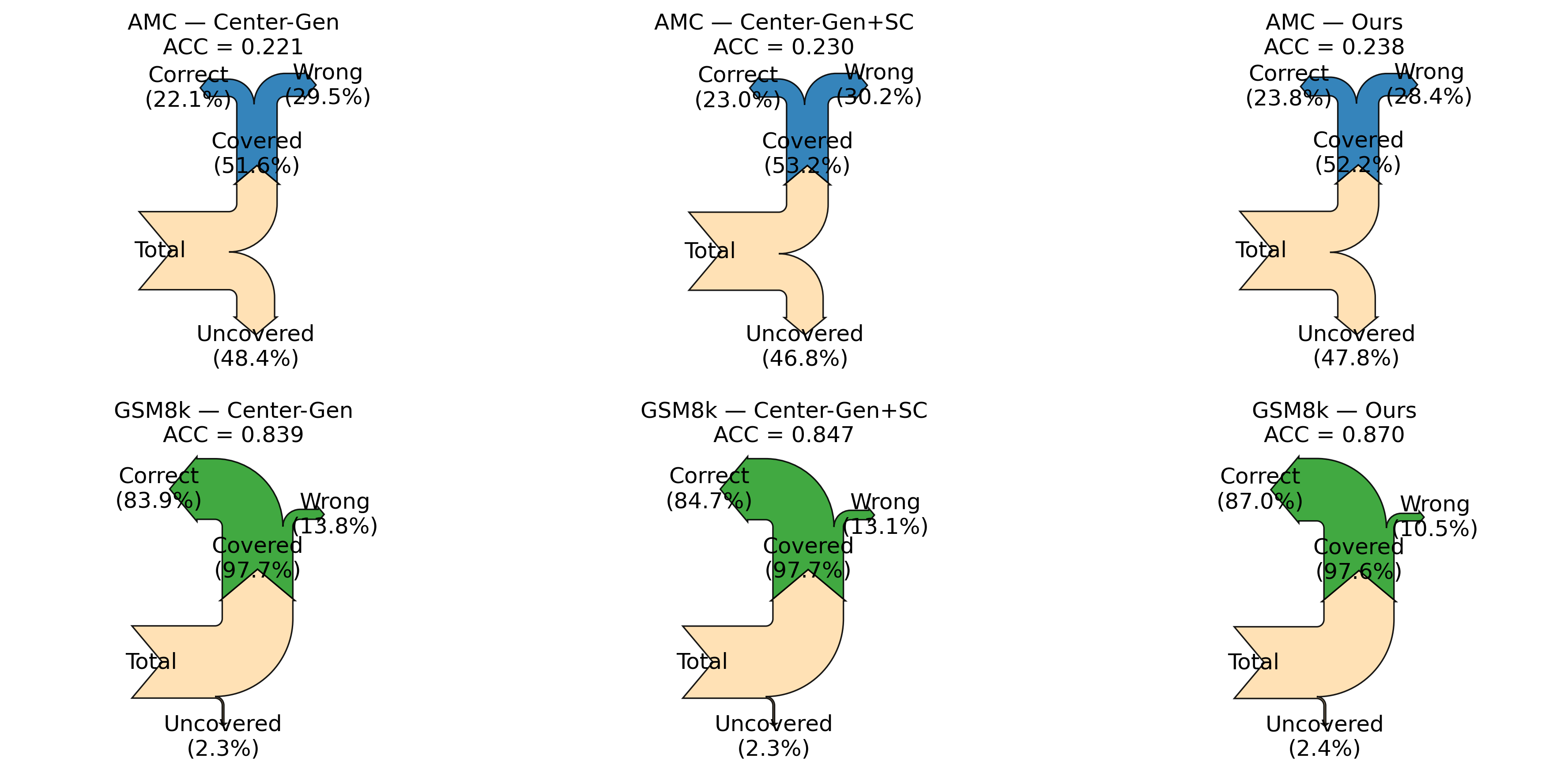}
    \caption{Sankey diagram illustrating performance on AMC and GSM8K under different central coordination strategies. Each flow decomposes accuracy into coverage and identification outcomes, showing how centralized selection more effectively converts diverse reasoning into correct solutions.}
    \label{fig:ab_sankey}
\end{figure}

\begin{table}[t]
\centering
\small
\begin{tabular}{lccc|ccc}
\toprule
\multirow{2}{*}{Model} & \multicolumn{3}{c|}{\textbf{LLaMA-8B (GSM8K)}} & \multicolumn{3}{c}{\textbf{Qwen-7B (GSM8K)}} \\
& Coverage & Identification & ACC & Coverage & Identification & ACC \\
\midrule
MAESTRO   & 0.9757 & 0.8919 & 0.8702 & 0.9886 & 0.9519 & 0.9410 \\
w/ CLPO   & 0.9773 & 0.9141 & 0.8933 & 0.9901 & 0.9607 & 0.9512 \\
\bottomrule
\end{tabular}
\caption{Comparison of coverage, identification, and accuracy (ACC) on GSM8K under two backbones.}
\label{tab:llm_backbone_acc_coverage}
\end{table}

\begin{figure}[h]
    \centering
    \includegraphics[width=0.5\linewidth]{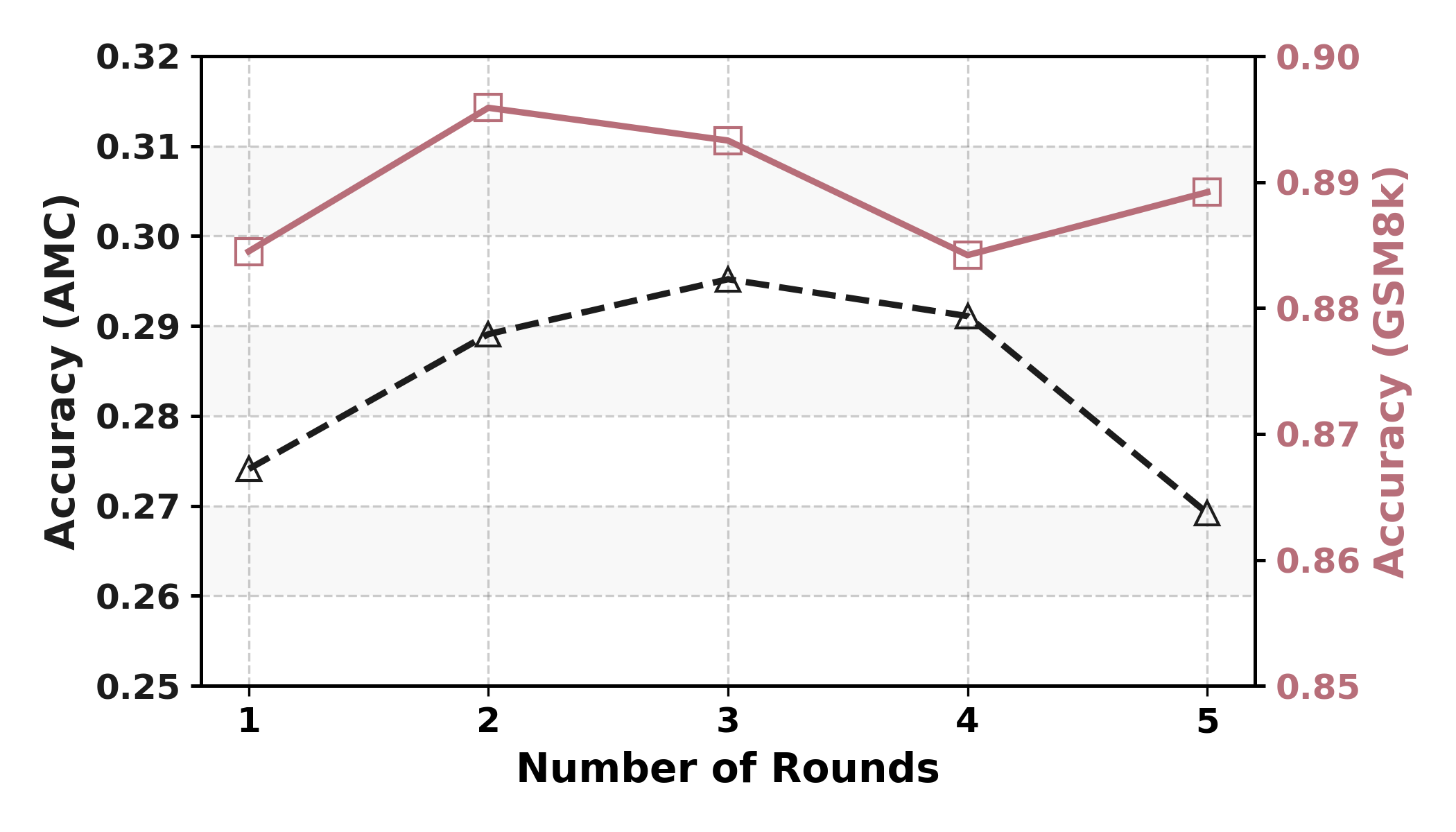}
    \caption{Effect of collaboration rounds on performance. Accuracy is reported for AMC and GSM8K. Performance improves with additional rounds up to a moderate level, then saturates or declines, highlighting the trade-off between evidence aggregation and bias amplification.}
    \label{fig:ab_rounds}
\end{figure}

\paragraph{Number of Rounds: balancing evidence aggregation and bias amplification.}
\label{app:exp_ab_agent_round}
Figure~\ref{fig:ab_rounds} shows that increasing the number of collaboration rounds improves performance at first, then saturates and may decline. GSM8K peaks around two rounds and AMC around three rounds. This pattern matches our coverage–identification decomposition. The first additional round injects public evidence through broadcast, which focuses subsequent exploration and lifts the identification probability \(q_t\) because the central policy compares candidates under a clearer hypothesis space. Coverage \(p_t\) changes little once the initial candidate pool is large, so early gains are mainly due to improved identification. Beyond the peak, returns diminish and can turn negative. Repeated conditioning on previous broadcasts reduces effective diversity and increases redundancy, which lowers the probability that new rounds add genuinely novel evidence. If an early broadcast is confidently wrong, later rounds tend to herd toward the same error, creating bias amplification that hurts \(q_t\). More rounds also introduce additional stochastic variance while consuming budget, which further limits net gains. Overall, a small number of rounds is most effective: two rounds on GSM8K and two to three rounds on AMC strike a good balance by converting collective coverage into reliable identification without over-conditioning the agents.

\paragraph{Sampling Depth per Agent: Coverage–Variance Trade-off.}
\label{app:exp_ab_sample_size}
We analyze how the number of samples drawn by each agent (\(K\)) affects collaborative performance. As shown in Figure~\ref{fig:number_agents_k} (b), increasing \(K\) from 2 to 3 improves accuracy on both AMC (from \(0.2369\) to \(0.2852\)) and GSM8K (from \(0.8853\) to \(0.8933\)). Beyond this point the gains saturate: GSM8K changes marginally at \(K{=}4\) (\(0.8936\)) and declines at \(K{=}5\) (\(0.8889\)), while AMC peaks at \(K{=}3\) and then drops to \(0.2811\) and \(0.2690\). This pattern reflects our exploration–synthesis decomposition. Increasing \(K\) initially raises the chance that at least one candidate is correct, which improves coverage \(p_t\). However, multi-sampling from the \emph{same} agent policy quickly becomes correlated and redundant, so the marginal gain in coverage diminishes. At the same time the candidate set grows and introduces more plausible distractors, which elevates the burden on the central selector and can depress identification \(q_t\). In practice, deeper per-agent sampling also inflates within-round variance because it relies on stochastic decoding from a single policy instance rather than diversifying across agents. Consequently, once coverage is near saturation, additional \(K\) contributes more noise than signal and identification becomes the limiting factor. Taken together with the agent-scaling results, these observations suggest a practical guideline: for a fixed budget, allocate capacity to increasing the \emph{number of agents} to diversify hypotheses, and keep \(K\) modest (three to four at most) so that the central synthesis can reliably convert collective coverage into higher identification.

\section{Declaration on the Use of Large Language Models}
In the preparation of this work, the authors used GPT-5 and GPT-4o for two specific purposes. First, GPT-5 was employed to polish the writing, improve clarity, and ensure grammatical correctness throughout the manuscript. Second, GPT-4o was used during dataset construction to assist in evaluating the quality of reasoning annotations. After using these tools, the authors reviewed and edited all content as needed and take full responsibility for the final version of the publication.

\end{document}